\pdfoutput=1

\documentclass[11pt]{article}

\usepackage{acl}
\usepackage{times}
\usepackage{latexsym}

\usepackage[T1]{fontenc}

\usepackage[utf8]{inputenc}

\usepackage{microtype}

\usepackage{inconsolata}

\usepackage{graphicx}
\usepackage{subcaption}
\usepackage{amsmath}
\usepackage{amssymb}
\usepackage{mathtools}
\usepackage{array}
\usepackage{xcolor}
\usepackage{tcolorbox}
\usepackage{lipsum} 
\usepackage{booktabs} 
\usepackage{bbding}
\usepackage[inkscapelatex=false]{svg}
\usepackage{hyperref}
\usepackage{bm}
\usepackage{fontawesome} 
\newcommand{\R}{\mathbb{R}}
\newcommand{\dt}{\Delta}
\newcommand{\A}{A}
\newcommand{\B}{B}
\newcommand{\C}{C}

\newcommand{\dA}{\overline{A}}
\newcommand{\dB}{\overline{B}}
\newcommand{\dC}{\overline{C}}

\newcommand{\dtAB}{(\dt, \A, \B)}

\title{Revealing and Mitigating the Local Pattern Shortcuts of Mamba}

\author{Wangjie You$^{1}$\thanks{\; Equal Contribution.}, Zecheng Tang$^1$\footnotemark[1], Juntao Li$^{1}$\thanks{\; Corresponding author.}, Lili Yao$^2$, Min Zhang$^{1}$ \\  
 $^{1}$School of Computer Science and Technology, Soochow University \\
 $^{2}$Machine learning platform department, Tencent  \\
 \texttt{\{wjyouuu,zctang\}@stu.suda.edu.cn} \\
 \texttt{\{ljt,minzhang\}@suda.edu.cn}; ~~~\texttt{liliyao@tencent.com}
 }

\begin{document}
\maketitle

\begin{abstract}
Large language models (LLMs) have advanced significantly due to the attention mechanism, but their quadratic complexity and linear memory demands limit their performance on long-context tasks.
Recently, researchers introduced Mamba, an advanced model built upon State Space Models~(SSMs) that offers linear complexity and constant memory.
Although Mamba is reported to match or surpass the performance of attention-based models, our analysis reveals a performance gap: Mamba excels in tasks that involve localized key information but faces challenges with tasks that require handling distributed key information. 
Our controlled experiments suggest that this inconsistency arises from Mamba's reliance on \textbf{local pattern shortcuts}, which enable the model to remember local key information within its limited memory but hinder its ability to retain more dispersed information.
Therefore, we introduce a global selection module into the Mamba model to address this issue.
Experiments on both existing and proposed synthetic tasks, as well as real-world tasks, demonstrate the effectiveness of our method.
Notably, with the introduction of only \textbf{4M} extra parameters, our approach enables the Mamba model~(130M) to achieve a significant improvement on tasks with distributed information, increasing its performance \textbf{from 0 to 80.54 points}.
\end{abstract}

\begin{center}
    \textbf{\textit{\faGithub~Code \& Data: \textcolor{violet}{ \url{https://github.com/ZetangForward/Global_Mamba.git}}}}
\end{center}

\section{Introduction}
In recent years, State Space Model~(SSM) has emerged as a promising successor to the attention-based models~\cite{vaswani2017attention} for long sequence modeling due to its linear computational complexity and constant memory requirements~\citep{gu2022efficiently,gupta2022diagonal,gu2022s4,smith2023s5}.
Different from the attention mechanism, which stores information for each token and performs pairwise computations between them, SSMs use a fixed-size state space to store history. 
This allows all computations to involve only the constant-sized state space.
Mamba~\citep{gu2023mamba}, built upon SSMs, is claimed to have achieved performance on par with, or even surpassing, that of attention-based models with the same parameters across language modeling and various synthetic tasks~\citep{pmlr-v235-dao24a,waleffe2024empirical}.

\begin{figure}[t]
    \centering
    \begin{subfigure}{0.48\textwidth}
        \centering
        \includegraphics[width=\linewidth]{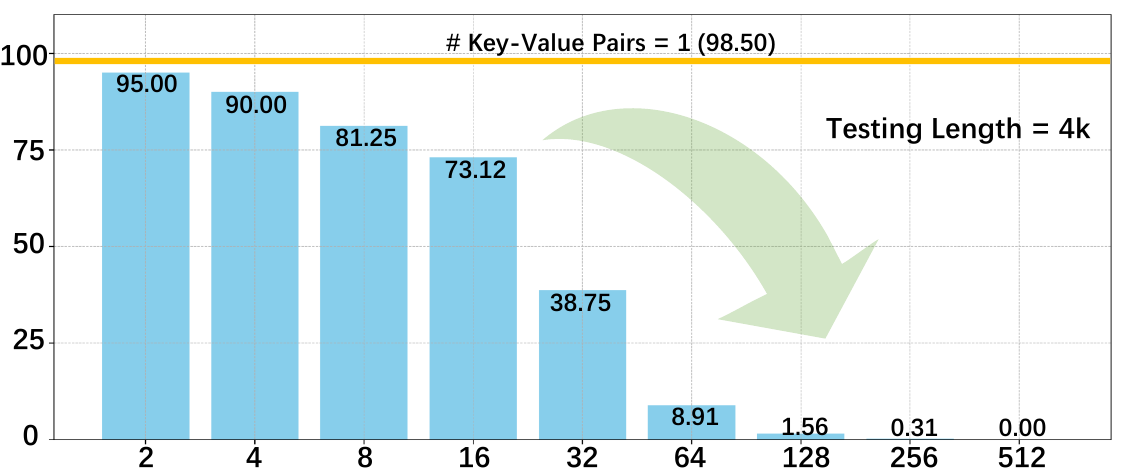}  
        \caption{ Performance as key-value pairs numbers increase.}
        \label{fig:sub1}
    \end{subfigure}
    
    \begin{subfigure}{0.48\textwidth}
        \centering
        \includegraphics[width=\linewidth]{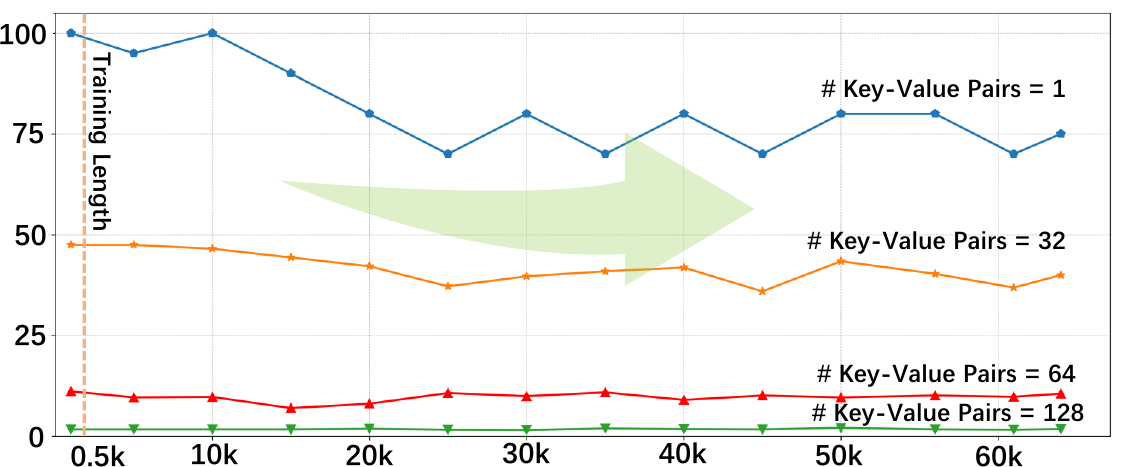}  
        \caption{Performance as testing length increases.}
        \label{fig:sub2}
    \end{subfigure}

    \caption{Mamba exhibits two distinct trends under different settings. The y-axis represents accuracy, while the x-axis in Fig.(a) shows the number of key-value pairs in the context with a testing length of 4K. In Fig.(b), the x-axis represents the testing length.}
    \label{fig: ppl_mqar}
\end{figure}

However, we observe an intriguing discrepancy in Mamba's performance on two settings of the \textsc{Mqar}\footnote{A synthetic task for testing a model's retrieval capability.} task: one requires the model to retrieve information from a local segment~(single Key-Value pair) within the context, while the other requires retrieving dispersed information~(multiple Key-Value pairs) from the context.
As shown in Fig~\ref{fig: ppl_mqar}(a), Mamba can effectively retrieve information from the local segment~(single Key-Value pair within a 4K context), even with a context length of up to 60K. 
However, in tasks that require extracting dispersed~(a few Key-Value pairs within a 4K context) or locally dense information~(a large number of Key-Value pairs within a 4K context), Mamba's performance is significantly affected and deteriorates sharply as the information density and dispersion increase.
Additionally, as shown in Fig~\ref{fig: ppl_mqar}(b), compared to the effects of information density and dispersion within the context, Mamba is relatively less affected by the context length.

In this paper, we conduct a controlled study to better understand the characteristics of Mamba under different context settings.
We start by analyzing Mamba's performance on different synthetic tasks, covering different information densities and dispersions.
Our findings reveal that Mamba relies on \textbf{local pattern shortcuts} to extract the desired information from the context, which manifests in two aspects: (1) \textbf{positional shortcuts}, i.e., Mamba tends to extract information from specific positions; and (2) \textbf{n-gram shortcuts}, i.e., Mamba tends to utilize specific high-frequent training templates to locate information.
These shortcuts make the model less robust to perturbed inputs, thereby further limiting Mamba's ability to generalize to complex tasks.
As a result, as previous works pointed out~\citep{amos2023never,park2024can,waleffe2024empirical}, Mamba performs well on tasks it has been trained on but struggles significantly on unseen tasks.
Furthermore, we explain the potential reasons for these shortcuts from two perspectives: the limited recurrent state size of the Mamba model and the constrained selectivity mechanism of SSMs.

To mitigate the aforementioned issues, we propose a global selection mechanism for the Mamba model. 
Specifically, we design an input-dependent global gating module for Mamba and observe significant improvements in its performance on complex synthetic tasks and the language modeling task.
Notably, by introducing \textbf{just 4M additional parameters} to the 130M-sized Mamba model, Mamba can achieve a breakthrough \textbf{from 0 to 80.54 points} on high information density synthetic tasks, significantly narrowing the performance gap between Mamba and attention-based models.

\section{Background}
 
\begin{figure*}[ht]
\centering
    \includegraphics[width=0.9\linewidth]{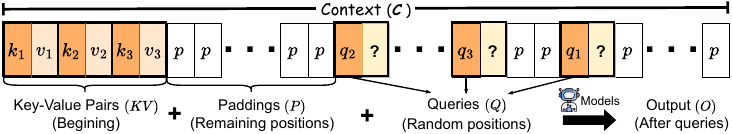}
    \caption{Illustration of \textsc{Mqar} Task.}
    \label{fig: mqar_task}
\end{figure*}

\subsection{State Space Model~(SSM)}
Structured state space sequence models (S4)~\cite{gu2021combining, gu2022s4,goel2022raw, ma2023mega, hasani2023liquid,smith2023s5}, represent a recent class of sequence models closely related to classical state space models.
These models are inspired by a specific continuous system that facilitates the mapping of a one-dimensional function or sequence \(x(t) \in \mathbb{R}\) to an output \(y(t) \in \mathbb{R}\) through an implicit latent state \(h(t) \in \mathbb{R}^N\).
Concretely, S4 models are characterized by four parameters \((\dt, \A, \B, \C)\), which define a sequence-to-sequence transformation as follows:

\begin{equation}
\begin{aligned}
\label{eq:state_space}
h^{\prime}(t) &= {\overline{A}} h_(t-1) + {\overline{B}} {x}(t) \\
y(t) &= {C} h(t)
\end{aligned}
\end{equation}
where $\dtAB$ are the discrete parameters, $\A \in \R^{N \times N}, \B \in \R^{N \times 1}, \C \in \R^{1 \times N}$, $\dA = f_A(\dt, \A)$, $\dB = f_B(\dt, \A, \B)$. 
Additionally, $f_A(\cdot), f_B(\cdot)$ are the pre-defined discretization functions.

\subsection{Selective State Space Model~(Mamba)} 
Selective State Space Model, as known as Mamba, is different from previous SSMs where the model's dynamics remain constant over time, it can efficiently update its hidden state based on the current input by introducing selective parameters.
Specifically, Mamba accomplishes this by employing specialized trainable linear layers that map the input to the matrices \(\overline{B}\), \(\overline{C}\), and the time step \(\Delta t\) for each processing step. 
Mamba conditions the discrete time-variant matrices dynamically based on the input as follows:
\begin{align}
    \begin{aligned}
         \dt_t = \tau &(S_{\dt} X_t), 
        \quad \dB_t = S_B X_t, \\
        & \quad \dC_t = (S_C X_t)^T, \\
    \end{aligned} \\
    \begin{aligned}
         \dA_t = \exp &(\A_t \dt_t) , 
        \quad \dB_t = \B_t \dt
    \end{aligned}
    \label{eq:reccurntRule}
\end{align}
where \(\dt t\) represents the discretization step, $\tau$ denotes the softplus function, and \(S_{\dt}\), \(S_B\), and \(S_C\) are linear transformation functions.
This enhancement empowers Mamba to execute more flexible sequence modeling, particularly for tasks demanding extensive historical information e.g., the Selective Copying task~\citep{arjovsky2016unitary} and the Induction Heads task~\citep{olsson2022context},  surpassing the performance of other SSMs.
Further discussions on other SSM variants and efficient model structures can be found in Appendix~\ref{appendix: related_models}.

\subsection{Multi-Query Associative Recall}
To make evaluations more controllable and eliminate the influence of the models' intrinsic knowledge, synthetic tasks are often employed~\cite {hsieh2024ruler}.
We conduct a further discussion of the synthetic tasks in Appendix~\ref{appendix: related_tasks}.
Among them, the Multi-Query Associative Recall~(\textsc{Mqar} is a widely adopted synthetic task for SSMs.
In \textsc{Mqar}, an input $x$ is structured as a sequence of bigrams representing key-value pairs, which are randomly drawn from a predefined dictionary.
Queries, i.e., the keys of key-value pairs, are then appended to this sequence, requiring the model to retrieve the corresponding value for the queried key.
As depicted in Fig.~\ref{fig: mqar_task}, formally, the input context \( \mathcal{C} = (c_0, \ldots, c_{N-1}) \) consists of $N$ tokens, where \( c_i \in \mathcal{V}\) and $\mathcal{V}$ is the vocabulary of the model.
We define \( N \) as the context length, representing the length of the input sequence.
The input sequence \( \mathcal{C} \) can be divided into three parts: key-value pairs \( \mathit{KV} \), queries \( \mathcal{Q} \), and padding tokens \( \mathcal{P} \).
The key-value pairs are \( \mathit{KV} = \{(k_1,v_1), (k_2,v_2), \ldots, (k_n,v_n)\} \), where \( n \) is the predefined number of key-value pairs. In the standard \textsc{Mqar} task, these key-value pairs are placed at the beginning of the sequence.
Queries are represented as \( \mathcal{Q} = \{q_1, q_2, \ldots, q_n\} \), where \( q_i = k_i \), and are inserted at random positions after the key-value pairs. Padding tokens are defined as \( \mathcal{P} = \mathcal{C} \setminus (\mathcal{Q} \cup \mathit{KV}) \), and they occupy the remaining positions in \( \mathcal{C} \), filled with random tokens.
The objective of the \textsc{Mqar} task is to predict:
\[
\mathcal{O}_i = \arg\max_{o_i \in \mathcal{V}} f_{\theta}(o_i \mid q_i,\mathit{KV},\mathcal{P}),
\]
where the model aims to output the most probable token \( \mathbf{o_i} \) from the vocabulary, given the padding tokens, key-value pairs, and the query \( \mathbf{q_i} \).
\textsc{Mqar} requires models to memorize key-value pairs in their hidden state, which presents a significant challenge for \textit{rnn-based} models, as they maintain a fixed-size state to handle all historical information.

\section{Analysis of Local Pattern Shortcuts in Mamba}
\label{sec:3}
Previous studies~\cite{gu2023mamba,arora2024zoology,arora2024simple} have shown that Mamba's success stems from its data-dependent features, where Mamba can dynamically gate the previous information based on the current input.
However, based on our preliminary study~(as shown in Fig.~\ref{fig: ppl_mqar}), we observe that Mamba performs poorly when the key information with the context becomes denser or more dispersed, regardless of various context lengths.
To discover the underlying reasons, we study the changes in the state space of the Mamba model during the inference process. 
In Sec.~\ref{subsec:lens_dynamic_selection_mechanism}, we first reformulate Mamba's process of assigning weights to each token into an attention-like matrix.
Then, we test the mamba model with the synthetic retrieval tasks and analyze the model's state space during the inference process.
Specifically, we utilize the 130M version of Mamba in all the experiments and design three different testing sets based on the \textsc{Mqar} task: (1) Positional Pattern Change, which alters the distribution of information in the context, moving beyond the previous \textsc{Mqar} task that places key information at the beginning~(Sec.~\ref{subsec:n_gram}); (2) N-gram Gathering, which controls the degree of aggregation of key information in the context by introducing more information within local segments, rather than solely testing the model's recall ability of single token~(Sec.~\ref{subsec:position}); and (3) Noise Injection, which adds noise tokens into the key information to perturb the model predictions, aiming to test the robustness of Mamba model~(Sec.~\ref{subsec:robust}).
\vspace{-1em}
\subsection{Reformulating Selection Process of Mamba into Attention-like Matrix}
\label{subsec:lens_dynamic_selection_mechanism}
Give the sequence $Y=\{y_1, y_2, \cdots, y_L\}$ that contains $L$ tokens, we leverage Eq.~\ref{eq:state_space} to calculate each $y_i$ and reformulate the calculation process into matrix multiplication, which can be written as:
\begin{equation}
\label{eq:attn_element}
Y = 
\begin{bmatrix}
y_1 \\ 
y_2 \\ 
\vdots \\ 
y_L 
\end{bmatrix} 
= \left[\begin{smallmatrix}
\alpha_{1,1} & \alpha_{1,2} & \cdots & \alpha_{1,L} \\
\alpha_{2,1} & \alpha_{2,2} & \cdots & \alpha_{2,L} \\
\vdots & \vdots & \ddots & \vdots \\
\alpha_{L,1} & \alpha_{L,2} & \cdots & \alpha_{L,L}
\end{smallmatrix}\right]
\begin{bmatrix}
x_1 \\ 
x_2 \\ 
\vdots \\ 
x_L 
\end{bmatrix},
\end{equation}
where $\alpha_{i,j}=C_i\left( \prod_{k=j+1}^{i} \bar{A}_k \right)\bar{B}_j~(i, j\in [1, L])$.

Then, we can transform Equ.~\ref{eq:attn_element} into an attention-like matrix by substituting and expanding $\alpha_{i,j}$: 
\begin{equation} 
\small
\label{eq:MambaAttn}
\begin{bmatrix}
    y_1 \\ y_2 \\ \vdots \\ y_L 
\end{bmatrix}
=
\underbrace{\left(\begin{smallmatrix}
    C_1 \bar{B}_1 & 0 & \cdots & 0 \\
    C_2 \bar{A}_2 \bar{B}_1 & C_2 \bar{B}_2 & \cdots & 0 \\
    \vdots & \vdots & \ddots & 0 \\
    C_L \prod_{k=2}^L \bar{A}_k \bar{B}_1 & C_L \prod_{k=3}^L \bar{A}_k \bar{B}_2 & \cdots & C_L \bar{B}_L
\end{smallmatrix}\right)}
\begin{bmatrix}
    x_1 \\ x_2 \\ \vdots \\ x_L
\end{bmatrix},
\end{equation}
where the portion enclosed by the \textbf{underbrace} represents the weights allocated by Mamba across the sequence, and we visualize this part to explore Mamba's state space.

\subsection{Positional Pattern Change}
\label{subsec:position}
\begin{figure}[t]
    \centering
    \includegraphics[width=\linewidth]{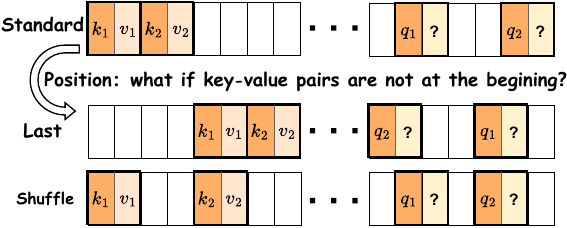}
    \caption{$\textsc{Mqar}$ task with different positional patterns.}
    \label{fig:position_task}
\end{figure}
\paragraph{Task Description}
As depicted in Fig.~\ref{fig:position_task}, in the standard \textsc{Mqar} task, all Key-Value pairs are placed at the beginning of the sequence.
This setup may lead the model to learn that it only requires ``remembering'' content from the initial portion of the sequence during training. 
To avoid such a fixed pattern of information distribution, we concentrate them at the end of the sequence~(Last) as well as disperse the key-value pairs from the beginning to arbitrary positions throughout the sequence~(Shuffle).
As depicted in Fig.~\ref{fig:position_task}, after adopting the aforementioned settings, i.e., Last and Shuffle, queries~($\mathcal{Q}$) are inserted at the random positions in the remaining padding sequence~($\mathcal{P}$).
Then, we train the Mamba model \textbf{separately} with training data containing three different positional patterns and evaluate it on \textbf{all} three test sets.
\paragraph{Results}

\begin{table}[t]
\centering
\small
\begin{tabular}{c|c|c|c}
\toprule
\bf Test \textbf{\textbackslash} Train  & Standard & Last &  Shuffle \\
\midrule
    Standard & 99.72 &  82.64  & 90.80 \\
    \midrule
    Last & 15.44 &  99.35  & 78.09 \\
    \midrule
    Shuffle & 22.37 &  54.08  &  80.98 \\
\bottomrule
\end{tabular}
\caption{Model performance on the Positional Pattern Change setting, where we report the prediction accuracy.}
\label{tab: position_res}
\end{table}

\begin{figure*}[t]
    \centering
    \begin{subfigure}{0.3\textwidth}
        \centering
        \includegraphics[width=\linewidth]{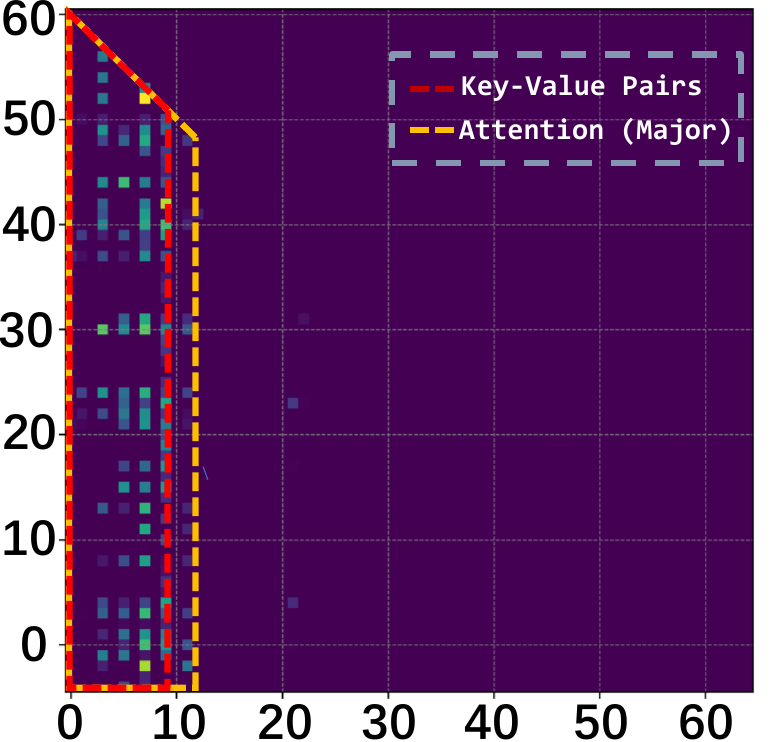}
        \caption{Standard-Standard}
        \label{fig:std-std}
    \end{subfigure}
    \begin{subfigure}{0.3\textwidth}
        \centering
        \includegraphics[width=\linewidth]{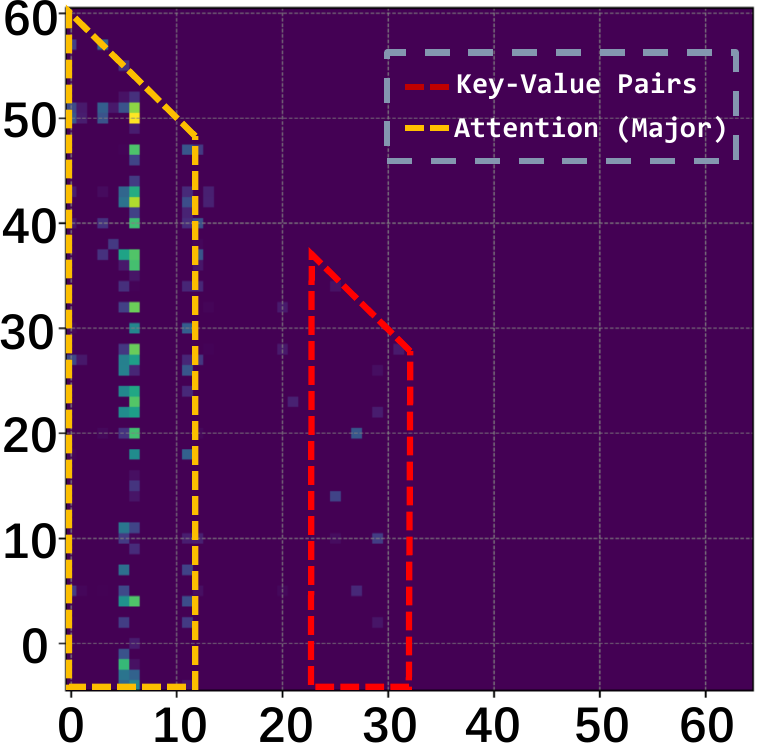}
        \caption{Standard-Last}
        \label{fig:std-last}
    \end{subfigure}
    \begin{subfigure}{0.3\textwidth}
        \centering
        \includegraphics[width=\linewidth]{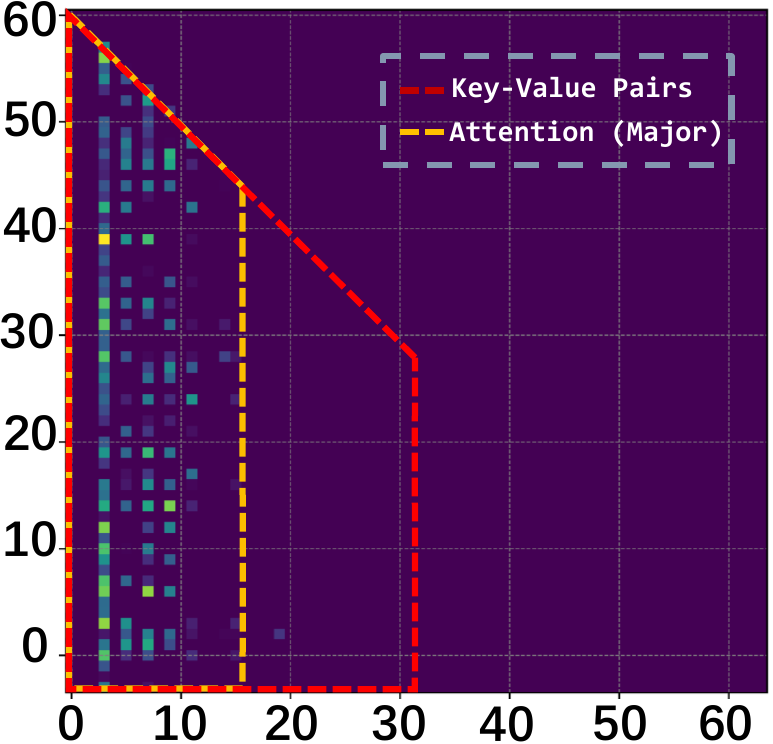}
        \caption{Standard-Shuffle}
        \label{fig:std-shuffle}
    \end{subfigure}

    \caption{The attention-like matrices of Mamba-130M that are trained on standard \textsc{Mqar} task and are tested on all three testing sets, i.e., Standard, Last, and Shuffle. 
    We plot the results of the 22nd layer of the model.
    Lighter colors indicate higher attention scores at specific positions.
    The red dashed line represents the location of the key-value pairs, while the yellow dashed line indicates where the model attends to the most.}
    \label{fig:mqar_attn}
\end{figure*}

As shown in Table~\ref{tab: position_res}, we can observe that when trained on the standard \textsc{Mqar} setup, Mamba achieved near-perfect accuracy on the in-domain test set~(standard), achieving 99.72 points.
In the other test settings, i.e., Last and Shuffle, Mamba only managed 15.44\% and 22.37\% accuracy, respectively.
However, when experimenting with the Mamba model on other settings, e.g., training with Shuffle data and testing with all three testing sets, Mamba performs consistently across all test settings, achieving or exceeding 50\% accuracy.
This finding reveals that \textbf{Mamba tends to learn information position shortcuts from the training data, allowing it to perform well on tasks with fixed template}, such as information extraction~(IE) tasks with fixed information positions. 
Notably, when Mamba is trained solely on the Last setting, its performance on the Shuffle test set is better than when trained on the Standard setting and tested on the Shuffle set, i.e., 54.08\% versus 22.37\%. 
This suggests that Mamba might \textbf{tend to memorize information at the beginning of the sequence}, making it difficult for the model to generalize to other settings when trained exclusively on the Standard \textsc{Mqar} training set.
where the portion enclosed by the \textbf{underbrace} represents the weights allocated by Mamba across the sequence, and we visualize this part to explore Mamba's state space. 
The visualized attention matrix of Mamba trained on the standard and test on all three test sets, as depicted in Fig.~\ref{fig:mqar_attn}, further supports this conclusion.
Despite the relocation of the key-value pairs, Mamba consistently attends to the beginning of the sequence, a behavior aligned with its training pattern but misaligned with the actual key-value pairs' positions.
This results in near-perfect performance in the standard setting but a noticeable decline in out-of-domain patterns i.e., Last and Shuffle.

\subsection{N-gram Gathering}
\label{subsec:n_gram}
\begin{figure}[t]
\centering
    \includegraphics[width=1 \linewidth]{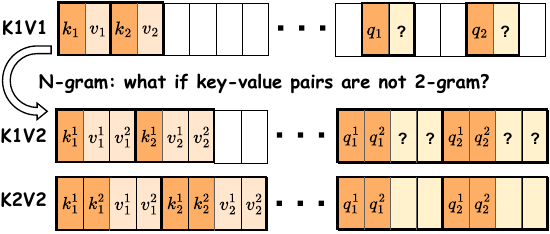}
    \caption{ Mqar task with different n-gram patterns.}
    \label{fig: ngram_task}
\end{figure}

\paragraph{Task Description}
In the standard \textsc{Mqar} task, models are required to predict the correct value given a key, where both the key and value are single tokens, i.e., $(k_i, v_i)\in \mathit{KV}$ and $|k_i|=|v_i|=1$.
We refer to this configuration as the \textit{K1V1} setting, primarily evaluating the model's 2-gram recall capability. 
However, addressing only 2-gram recall is insufficient, as the amount of information to be recalled is minimal, and learning specific 2-gram pairings is relatively easy. 
Moreover, most real-world entities involve multiple tokens, and testing 2-gram capability alone fails to reflect performance on other tasks. 
Therefore, to increase the amount of information to be recalled, we propose the N-gram Gathering setting.
As shown in Fig.~\ref{fig: ngram_task}, we increase the number of tokens required for recall in both the Key and Value portions, i.e., $|k_i|=N, |v_i|=M~(N>1, M>1)$
We refer to such a configuration as the \textit{KNVM} setting.
Specifically, we set up three data configurations: $\mathit{K1V1}, \mathit{K1V2}, \mathit{K2V2}$.
We train the model on each of these settings \textbf{separately} and then evaluate it across \textbf{all} three testing sets.

\paragraph{Results}

As shown in Table~\ref{tab: n_gram_res}, Mamba exhibits strong performance when the number of key-value tokens in the training set matches or exceeds those in the test set, e.g., training with $\mathit{K1V2}$ and testing with $\mathit{K1V1}$. 
However, its performance deteriorates significantly when the number of value tokens in the test set surpasses those encountered during training, e.g., training with $\mathit{K1V1}$ and testing with $\mathit{K2V2}$. 
This indicates that Mamba tends to learn simple patterns in tasks, e.g., learning to respond based on the single anchor token in $\mathit{K1V1}$, which hinders its generalization ability on other complex tasks.
Furthermore, Mamba's success would not be due to true n-gram recall but rather an \textbf{over-reliance on the structural cues provided by the special characters or templates}.
We refer to this phenomenon as the n-gram shortcut of Mamba.

\begin{table}[t]
\small
\centering
\begin{tabular}{c|c|c|c}

\toprule
\bf Test \textbf{\textbackslash} Train   & $\mathit{K1V1}$ & $\mathit{K1V2}$  &  $\mathit{K2V2}$ \\
\midrule
    $\mathit{K1V1}$  &  99.95 &  99.81  & 66.82 \\
    \midrule
    $\mathit{K1V2}$  &  0.00 &  99.96  & 96.66 \\
    \midrule
    $\mathit{K2V2}$  &  0.00 &  99.90  & 99.98 \\
\bottomrule
\end{tabular}
\caption{Model performance on the N-gram Gathering setting, where we report the prediction accuracy.}
\label{tab: n_gram_res}
\end{table}

\subsection{Noise Injection}
\label{subsec:robust}

\begin{figure}[t]
\centering
    \includegraphics[width=1 \linewidth]{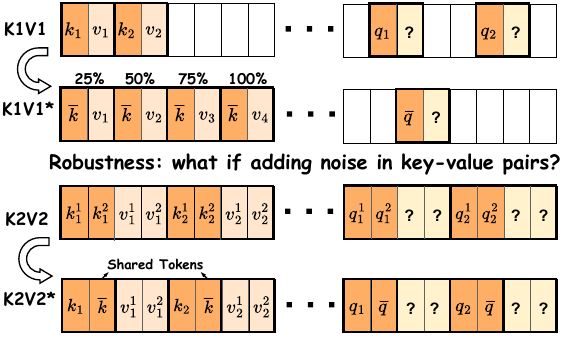}
    \caption{
    \textsc{Mqar} task with the injection of noise.
    *~denotes the robustness setting. 
    }
    \label{fig: robustness_task}
\end{figure}

\paragraph{Task Description}
We further explore Mamba's robustness, specifically its ability to generalize beyond the shortcuts mentioned above.
As shown in Fig.~\ref{fig: robustness_task}, we place $n$ sets of Key-Value pairs at the beginning of the sequence and divide these Key-Value pairs into four regions: $\mathit{KV} = \{(k_1, v_1), (k_2, v_2), (k_3, v_3), (k_4, v_4)\}$. 
Then, two settings are adopted: $\mathit{K1V1}^{*}$ and $\mathit{K2V2}^{*}$.
In $\mathit{K1V1}^{*}$, all the keys $k_i$ are identical, i.e., $k_1 = k_2 = k_3 = k_4$, and we utilize symbol $\overline{k_i}$ to denote those tokens.
In $\mathit{K2V2}^{*}$, the last token of $k_i$ are identical.
All the values $v_i$ are different in the above two settings.
Then, we let the model predict the corresponding value by providing $k_i$.

\paragraph{Results}
As shown in Fig.~\ref{fig: robustness_task}(a), in terms of the $\mathit{K1V1}^{*}$ setting, we observe that the model's performance during testing closely aligns with the training patterns. 
Specifically, when all tokens are placed at the beginning of the sequence during training~(Standard \textsc{Mqar}), the model tends to retrieve information from the front, primarily focusing on the first 25\% of the sequence. 
Conversely, if all key information is positioned toward the end of the sequence during training~(Last \textsc{Mqar}), the model tends to retrieve information from the latter parts of the sequence, concentrating mainly on the 75\% to 100\% regions.
This behavior highlights Mamba's \textbf{over-fitting to the positional patterns learned during training, which may lead to a reliance on positional shortcuts rather than recall capability.}
Besides, as shown in Fig.~\ref{fig: robustness_task}(b), we can observe that as the number of Key-Value pairs increases, Mamba maintains an accuracy exceeding 90\% in the presence of noise when there are four Key-Value pairs. 
However, as the amount of noise increases, Mamba's performance declines sharply.
This indicates that Mamba relies on the partial high-frequent information within the keys for its predictions.
Therefore, when noise overwhelms this critical information, the model's performance declines dramatically.

\section{Mitigating the Shortcuts of Mamba}
\subsection{Key to Selectivity of Mamba}
For \textit{attention-based} models, the recurrent state grows with the length of the sequence, enabling perfect recall accuracy but at the cost of efficiency.
In contrast, \textit{rnn-based} models maintain a fixed recurrent state size, making it critical to optimize the use of their limited memory resources.
Mamba stands out by efficiently balancing the memory-recall trade-off through its data-dependent design.
By introducing selective parameters, it updates its hidden state i.e.,  which information is retained or discarded based on the current input.
This mechanism is formulated as follows:
\begin{equation}
h_t = \bar{A} h_{t-1} + \bar{B} x_t = e^{A \Delta t} h_{t-1} + \Delta t B x_t,
\label{fig: mamba_h}
\end{equation}
where $\Delta_t = \tau$$(S_{\Delta} X_t)$. 
$\Delta_t$ is the key to the selectivity of Mamba as it simultaneously determines the selective nature of the matrices $\bar{A}$ and $\bar{B}$ through the linear transformation function, which ultimately decides the update of the recurrent state.

$\Delta t$ is generated from the current input via the short convolution function, followed by two linear transformations in the original Mamba.
The original goal of this function is to extract local features and establish local contextual relationships between tokens before entering the SSM module.
However, the short convolution establishes contextual relationships between tokens within a limited scope.
In scenarios where information is distributed across distant positions, this reliance can lead to shortcuts, causing the model to overly depend on simple features learned during training (e.g., positional cues) rather than capturing more meaningful dependencies.

\begin{figure}[t]
\centering
    \includegraphics[width=1 \linewidth]{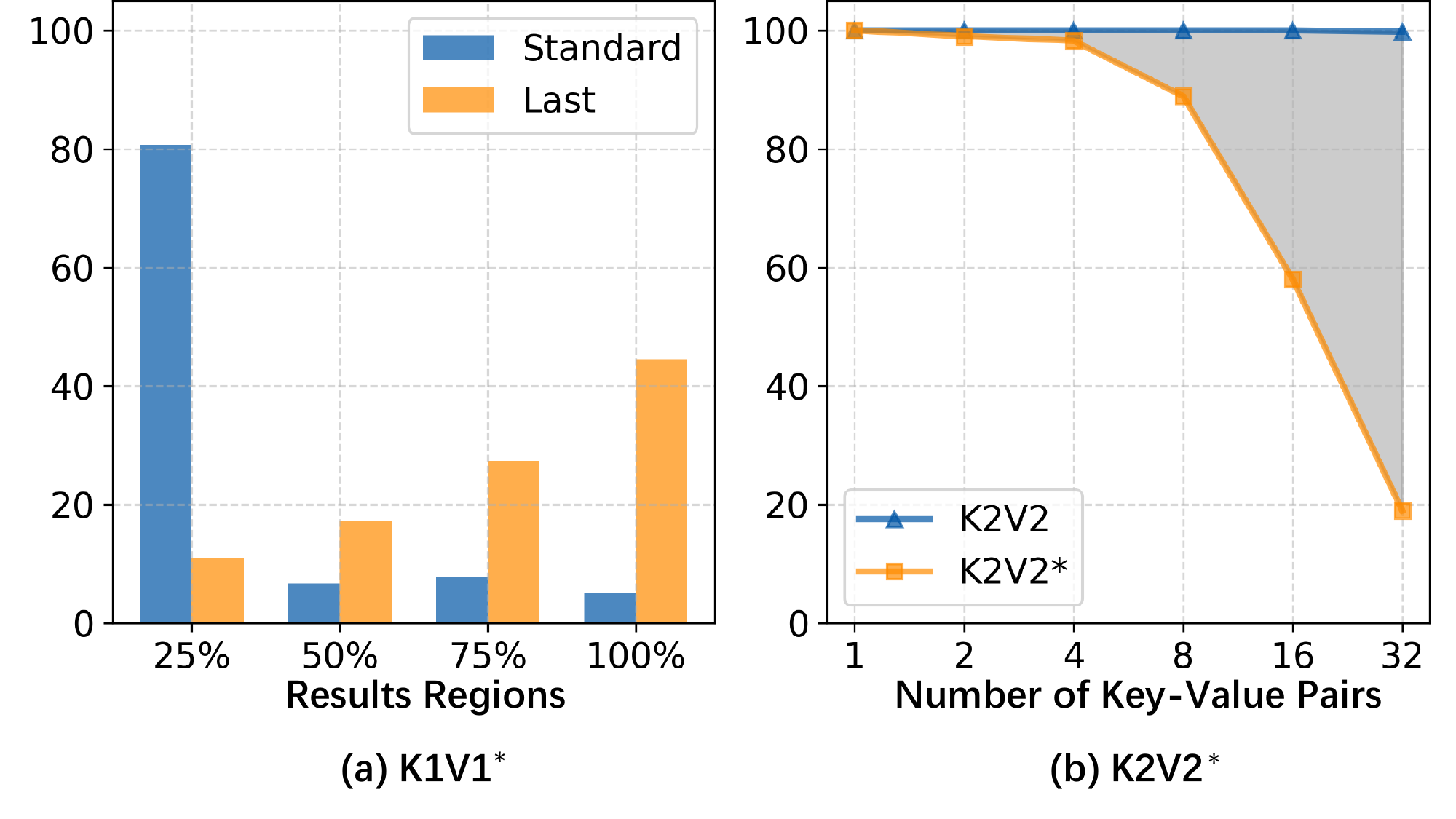}
    \caption{ (a) Results of $\mathit{K1V1}^{*}$ setting after training with different position modes~(Standard \textsc{Mqar} and Last \textsc{Mqar}), where the x-axis represents the ratios of regions. (b) Results of $\mathit{K2V2}^{*}$ setting, where
    the model is trained with standard $\mathit{K2V2}$ data and the x-axis represents the number of key-value pairs.}
    \label{fig: robustness_res}
\end{figure}

\subsection{Incorperating Global Selection Mechanism}
 To address this issue, we propose incorporating more global information into \( \Delta t \) through a fine-grained global selection mechanism.
 Specifically, we introduce a long convolution module to model distant context and integrate its output into the original \( \Delta t \). 
 This process is formulated as follows:
    \begin{equation}
    \begin{aligned}
    \dt_t &= \mathbf{W}_2 \cdot \sigma\left( \mathbf{W}_1 \cdot \text{Conv}_{\text{short}}(\mathbf{X}_t) \right) \\
    &\odot \sigma\left( \text{Conv}_{\text{long}}(\mathbf{X}_t) \right)
    \end{aligned}
    \end{equation}
where \text{Conv} denotes the convolution operation, \(\mathbf{W}\) represents the linear transformation operations, \(\sigma\) is the nonlinear activation function, specifically the Silu~\citep{ramachandran2017swish} activation function, and \(\odot\) is the element-wise multiplication operation.
The result of the long convolution serves as a gating mechanism, enabling global selection on the original \(\Delta_t\).
By introducing this global information module, Mamba's selectivity is better aligned with global decision-making rather than being constrained by local patterns.

\begin{table*}[t]
\centering
\resizebox{\textwidth}{!}{
\renewcommand{\arraystretch}{1} 
\setlength{\tabcolsep}{10pt} 
\begin{tabular}{l l | cccccc}
\toprule
\textbf{Models} & \textbf{Scale} & \textbf{Shuffle} & \textbf{Std-Last} & \textbf{Std-Shuffle} & \textbf{K2V2} & \textbf{K2V2-Robustness} & \textbf{K4V8-Shuffle} \\ 
\midrule
\textbf{Pythia }~\citep{biderman2023pythia} & \textit{133m} & 99.82 & 93.75 & 94.31 & 99.99 & 99.99 & 99.99 \\ 

\midrule

\textbf{Hyena}~\citep{poli2023hyena} & \textit{153m} & \XSolidBrush & \XSolidBrush  &  \XSolidBrush & 77.62 & 65.92 & 22.51  \\

\textbf{RWKV }~\citep{peng2023rwkv} & \textit{153m} & \XSolidBrush & \XSolidBrush  & \XSolidBrush  & 85.99 & 72.62 & 6.57 \\

\textbf{Mamba }~\citep{gu2023mamba} & \textit{129m}& 80.98 & 15.44 & 22.37 & 99.98 & 66.01 & \XSolidBrush \\ 
\hspace{0.8em} \textbf{w/}    2$\times$\textit{State Size} & \textit{130m} & 88.57 & 40.22 & 31.88 & 99.84 & 78.90 & \XSolidBrush \\ 
\hspace{0.8em} \textbf{w/}     4$\times$\textit{State Size} & \textit{134m} & 96.92 & 35.89 & 32.88 & 99.84 & 57.11 & \XSolidBrush \\ 
\hspace{0.8em} \textbf{w/}    \textit{Global Selection} & \textit{133m} & 90.45 & 41.97 & 35.73 & 99.06 & 81.46 & 80.54 \\ 
\bottomrule
\end{tabular}
}
\caption{
Performance of models on variations of \textsc{Mqar} tasks.
\XSolidBrush denotes that the model fails with this setting with an accuracy lower than 5\%.
\XSolidBrush indicates that the model fails in this setting, with an accuracy lower than 5\%. \textit{State Size} refers to the Mamba model with an increased state space size, which was originally set to 16.
}
\label{table:main_exp}
\end{table*}

\section{Experiment}
In this section, we conduct more comprehensive experiments across different models with varying foundational mechanisms, i.e., \textit{attention-based} and \textit{rnn-based}, to explore whether they exhibit similar behavior as observed in Mamba.
Meanwhile, we evaluate whether our proposed prototype, which incorporates global selection, can mitigate Mamba's local pattern shortcut issue and assess its performance on real-world downstream tasks.

\subsection{Experimential Settings}
\paragraph{Datasets}
We selected tasks where the original Mamba exhibited clear shortcuts and performed poorly.
The settings are the same as outlined in Sec.~\ref{sec:3}.
For the Shuffle and $\mathit{K2V2}$ settings, models are trained and tested on the same corresponding setting.
We evaluate the model's in-domain capabilities in scenarios with increased information density and dispersion.
In contrast, the Last and Std-Shuffle tasks test the model's out-of-domain performance, where the testing mode is inconsistent with the training one, i.e., training on the standard position pattern and testing on the Last and Shuffle pattern.
In the $\mathit{K2V2}$-Robustness and $\mathit{K4V8}$-Shuffle settings, we evaluate whether the models can mitigate the influence of noise as well as handle tasks with both high information density and divergence.

For downstream tasks, we evaluate whether the global selection strategy negatively impacts or enhances downstream task performance.
Following previous works, we report perplexity on Wikitext (Wiki.)~\citep{paine2016fast} and additional tasks such as PIQA~\citep{bisk2020piqa}, HellaSwag~\citep{zellers2019hellaswag}, and Winogrande~\citep{sakaguchi2021winogrande}.
We use the lm-evaluation-harness~\citep{eval2021harness} for evaluation.

\paragraph{Baselines} We adopt one representative \textit{attention-based} model, Pythia~\citep{biderman2023pythia}, along with two \textit{rnn-based} models i.e., Hyena~\citep{poli2023hyena} and RWKV~\citep{peng2023rwkv} at allied scale.
We aim to investigate whether models with varying architectures would exhibit similar local pattern shortcuts as Mamba.
For Mamba models, intuitively, increasing the state size allows the model to store more historical information, reducing the compression of past data and decreasing the likelihood of the model relying on shortcuts.
Therefore, we propose a simple baseline by increasing Mamba's \text{SSM-State} size as baselines.
All models are trained on the same set according to different settings.
More detailed data configurations, model training hyperparameters, and evaluation settings can be found in Appendix~\ref{appendix: Models},~\ref{appendix: ana_exp}, and ~\ref{appendix: downstream}.

\subsection{Main Result}
\subsubsection{Synthetic Tasks}
\paragraph{Attention-based Model}

As depicted in Tab.~\ref{table:main_exp}, Pythia performs nearly perfectly across all tasks, as expected, given that \textit{attention-based} models excel at accurately identifying the corresponding correct answers through pairwise token interactions.
Unlike models constrained by state size, \textit{attention-based} models inherently capture long-range dependencies by attending to the entire input sequence.

\paragraph{Other RNN-based Models}
In contrast to \textit{attention-based} models, \textit{rnn-based} models perform significantly worse across all tasks, reflecting their inherent limitations on the \textsc{Mqar} task requiring dynamic state adaptation due to their fixed state space size.
However, neither Hyena nor RWKV exhibits the positional pattern shortcut observed in Mamba, as they do not show an obvious gap between in-domain and out-of-domain performance.
Additionally, despite trailing Mamba in the standard $\mathit{K2V2}$ tasks, both models show comparable performance against Mamba in the $\mathit{K2V2}$-Robustness setting. 
This suggests that Mamba is more susceptible to noise, likely due to its reliance on n-gram shortcuts. These observations further suggest that Mamba's shortcut behavior is closely tied to its unique selective mechanism, which prioritizes local patterns over the global context due to the constrained $\dt_t$.

\paragraph{Mamba} 
With regard to the strategies posed on Mamba, they both yield improvements across most tasks.
Notably, although increasing the recurrent state size generally enhances performance, the improvement is not consistently proportional to the size increase.
For instance, in the $\mathit{K2V2}$-Robustness task, a recurrent state size of \(4\) times compared to the original one led to a performance drop to \(57.11\%\), lower than the \( 2\) times of state size and even below the original Mamba model's performance of \(66.01\%\).
This indicates that merely enlarging the recurrent state size is not a sufficient solution.
While a larger recurrent state size enables models to store more historical information, it may also introduce irrelevant information if the model cannot effectively update the recurrent state space through an active selectivity mechanism.

On the other hand, making more effective updates to the recurrent state space by incorporating global information into the selectivity module, i.e., Global Selection, led to more consistent improvements across the board without introducing a significant increase in parameter count. 
Notably, Global Selection outperforms other configurations in out-of-domain settings; however, it relatively lags behind on in-domain tasks i.e., Shuffle and $\mathit{K2V2}$.
This suggests that integrating distant information into the decision-making process effectively mitigates the model's reliance on local shortcuts; similarly, this mechanism may explain its disadvantage on in-domain tasks, where shortcuts provide a performance advantage.
In the $\mathit{K2V2}$-Robustness task, the Global Selection method achieved a score of \(81.46\%\), significantly outperforming all other configurations, demonstrating its effectiveness in mitigating the impact of noise.
Furthermore, in the $\mathit{K4V8}$-Shuffle task, which posed substantial difficulties for the original Mamba model, the Global Selection method scored \(80.54\%\), highlighting its superior capability in handling tasks with high information density and divergence.

These findings suggest that for \textit{rnn-based} models with a fixed recurrent state size, the key to achieving consistent improvements lies in effectively maintaining and accurately updating the recurrent state rather than merely increasing its capacity to store more historical information.

\begin{table}[t]
\small
\centering
\begin{tabular}{l|c|c|c|c}

\toprule
\bf Models   & \textbf{Wiki.} & \textbf{PIQA}  &  \textbf{Hella.} & \textbf{Wino.} \\
     & ppl ${\downarrow}$  & acc ${\uparrow}$  & acc ${\uparrow}$ & acc ${\uparrow}$  \\
\midrule
    Pythia                         & 40.94 & 61.21 & 27.90 & 51.14\\
    \midrule
    Mamba                          & 40.46 & 62.13 & 28.70 & 52.25 \\
     \hspace{0.5em} w/ \textit{GS}   & \textbf{38.92} & \textbf{62.62} & \textbf{28.79} & \textbf{52.41} \\

\bottomrule
\end{tabular}
\caption{ The results of models at 130m scale on downstream tasks, \textit{GS} denotes the global selection strategy }
\label{tab: downstream_task}
\vspace{-15pt}
\end{table}

\subsection{Downstream Task}
While the synthetic tasks provide valuable insights into how Global Selection mitigates shortcut behavior in Mamba, it is critical to assess how these improvements transform into real-world downstream tasks.
As shown in Table~\ref{tab: downstream_task},
Mamba has already demonstrated notable success on downstream tasks, surpassing the performance of the \textit{attention-based} baseline model, Pythia. 
The introduction of Global Selection further enhances Mamba's performance in language modeling and commonsense reasoning tasks.
The Global Selection strategy consistently achieves the best results across all evaluated tasks, underscoring its effectiveness across different domains.
The most significant improvements are observed in the language modeling task, where the perplexity is reduced by 1.54 when compared to the original Mamba, underscoring the model's enhanced ability to handle long-range dependencies and avoid over-reliance on local patterns.

\section{Conclusion}
In this work, we extend the \textsc{Mqar} task to investigate Mamba's underlying behavior.
Our controlled experiments reveal Mamba’s reliance on local pattern shortcuts.
To address this, we introduced a fine-granted selection mechanism into the Mamba model by incorporating global information into the decision-making factor $\dt_t$.
Experiments on both existing and proposed synthetic tasks, as well as real-world tasks, demonstrate the
effectiveness of our method.
Our findings suggest that for \textit{rnn-based} models with a fixed recurrent state size, efficiently utilizing the available state space is far more critical than simply increasing its capacity.

\section*{Limitation}
Our approach is primarily focused on experimentally analyzing the shortcut phenomenon in Mamba; further exploration of theoretical insights is needed.
The proposed method aims to address the shortcut issues observed in Mamba on synthetic tasks.
However, further improving performance on downstream tasks may require additional adaptations and comprehensive testing across a wider range of models with varying scales.
While preliminary results indicate that similar shortcuts were not present in other \textit{rnn-based} models, further validation is needed to determine whether our findings generalize across diverse architectures.

\bibliography{main}

\appendix

\section{Related Work} 
\label{appendix: related_tasks}
\subsection{Synthetic Task} 
Synthetic tasks have played a crucial role in advancing language modeling, serving as controlled, simplified, and purposefully constructed benchmarks for assessing specific model capabilities. By providing a controlled environment, these tasks enable researchers to isolate particular challenges and evaluate how models handle them, offering deeper insights into their underlying mechanisms.
Such tasks not only facilitate the identification of strengths and weaknesses in language models but also drive innovation and optimization in model architecture and training strategies.
In contrast to more complex and realistic benchmarks, synthetic tasks offer greater flexibility in regulating various factors like sequence length, task complexity, and input structure.
This flexibility makes them invaluable for probing specific behaviors of language models without interference from large-scale parametric knowledge or the unpredictability of real-world data.
As a result, they serve as ideal testing grounds for understanding how models manage tasks that require attention over long sequences or the processing of intricate patterns.

Numerous synthetic tasks have been developed to test different dimensions of model performance. 
For example, copying and selective copying tasks~\citep{jing2019gated} evaluate a model’s memory retention and replication abilities, while tasks involving induction heads~\citep{olsson2022context} examine its capacity to infer relationships in context. Others, such as passkey retrieval~\citep{landmark}, needle-in-a-haystack~\citep{needle}, and associative recall~\citep{graves2014neural,ba2016using}, have been instrumental in testing how well large language models (LLMs) handle long-range dependencies, particularly in extremely long sequence contexts.
In addition, ~\citeauthor{hsieh2024ruler} propose a novel synthetic benchmark \textsc{Ruler} to evaluate long-context language models.

The \textit{associative recall} (\textsc{AR}) task, inspired by psychological models of how humans associate and retrieve information, has been a focal point of early neural network research aimed at developing systems capable of associative recall.
With the advent of large language models, many researchers have argued that the ability of LLMs to perform \textit{in-context} learning is, at least in part, attributable to the associative recall capabilities embedded in attention mechanisms~\citep{elhage2021mathematical, olsson2022context}.
More recently, several notable recurrent neural network architectures have been evaluated using synthetic versions of the associative recall task ~\citep[inter alia]{graves2014neural,ba2016using,zhang2017learning}.

Our work builds upon and extends one of these tasks, specifically the Multi-Query Associative Recall (\textsc{Mqar})~\citep{arora2024zoology}.
In contrast to the standard associative recall task, \textsc{Mqar} is designed to more closely resemble the complexities of natural language processing.
It introduces multiple key-value pairs and challenges models to retrieve the correct associations despite distractors, thereby offering a more rigorous test of a model's ability to manage selective attention and long-range dependencies within dynamic, varied input structures.
This task pushes the boundaries of synthetic evaluations by bridging the gap between controlled experimental setups and the intricate nature of real-world language tasks.
In this work, we extended the \textsc{Mqar} task by introducing variations in positional and n-gram patterns to investigate Mamba's underlying behavior.
This analysis framework can also be adapted for evaluating other models, providing a broader tool for identifying similar issues.

\subsection{Efficient Model Architecture}
\label{appendix: related_models}

Efficient model architectures (\textit{e.g., State-Space Models and Linear Attentions }) have become increasingly popular due to their ability to scale to long sequences while maintaining competitive performance. In this section, we focus on several key models that illustrate different strategies for improving efficiency.

\textbf{S4} \citep{gu2021combining} introduced a novel class of sequence models designed to capture long-range dependencies using a state-space model (SSM) framework, typically formulated in continuous time. S4 introduces three key mechanisms to achieve this: (1) the Higher-Order Polynomial Projection Operator (HiPPO) \citep{gu2020hippo}, which efficiently memorizes signal history by operating on state and input transition matrices, (2) a diagonal plus low-rank (DPLR) parametrization that stabilizes the SSM matrix (A) by adding a low-rank correction, ensuring both diagonalizability and stability, and (3) efficient kernel computation through Fast Fourier Transforms (FFT) and inverse FFTs, reducing the overall complexity to $\mathcal{O}(N \log N)$. By leveraging these innovations, S4 significantly improves the handling of long-range dependencies, offering a more scalable alternative to traditional models.
The SSM parameters in S4~\citep{gu2022s4} and S5~\citep{smith2023s5} are fixed after training, resulting in data-independent configurations that significantly limit the overall expressivity of models. 
In contrast, Mamba~\citep{gu2023mamba} addresses this limitation by introducing data-dependent parameters for S4.

\textbf{S5} \citep{smith2023s5} shifted the formulation from a single-input single-output (SISO) to a multi-input multi-output (MIMO), reducing the effective state dimension while retaining computational scalability, making it a key advancement in the structured SSM framework.
S5 also independently discovered the diagonal SSM approximation and became the first S4 variant to implement recurrent computation via parallel scan. 

\textbf{H3} \citep{fu2023hungry} was developed by combining S4 with linear attention \citep{kaul2020linear}, generalizing this formulation to more complex recurrent structures. This model is the first to extend linear attention to broader recurrences, which form the foundation for later architectures. By integrating the strengths of SSMs and linear attention, H3 effectively balances scalability and performance in long-sequence processing tasks.

\textbf{Hyena} \citep{poli2023hyena} was designed to close the perplexity gap between attention-based Transformers and sub-quadratic alternatives. While traditional attention mechanisms face quadratic complexity with increasing sequence length, Hyena overcomes this by using implicitly parameterized long convolutions and data-gating, achieving sub-quadratic complexity. The model also shows that previous sub-quadratic attention approaches, such as those based on low-rank or sparse approximations, often still rely on dense attention layers to reach Transformer-level performance.

\textbf{Retnet} \citep{sun2023retentive} is an efficient model for long-range sequence modeling, built on linear attention with a simplified internal structure that reduces the state dimension to one. Its recurrence mechanism can be viewed as a specific case of linear state space models (SSMs). RetNet introduces the retention mechanism, which supports three computation modes: parallel, recurrent, and chunkwise recurrent. These modes enable parallel training, low-cost O(1) inference for faster decoding, and efficient long-sequence modeling with linear complexity by processing chunks in parallel and summarizing them recurrently.

\textbf{RWKV} \citep{peng2023rwkv} is a recent RNN architecture designed specifically for language modeling, featuring a linear attention approximation mechanism called the "WKV" mechanism. 
RWKV utilizes linear time-invariant recurrences and can be viewed as the ratio of two state space models.
Unlike traditional Transformers, which have quadratic complexity in terms of computation and memory, RWKV offers linear scalability, combining the efficiency of RNNs with the performance of Transformers. While RWKV is presented as a hybrid model of RNNs and Transformers, it primarily relies on linear attention and lacks the recurrent properties of traditional RNNs, making it more similar to attention-based models.

\textbf{Based} \citep{arora2024simple} is a simple yet flexible architecture that integrates linear attention with sliding window mechanisms. By varying the window size and the feature dimension of the linear attention layer, Based can navigate the Pareto frontier of the recall-memory tradeoff. This allows it to effectively achieve full attention-like quality on one end while providing a compact state size for memory-efficient alternatives on the other.

\textbf{GLA} \citep{yang2024gated} introduces an efficient training algorithm for linear attention Transformers that integrates data-dependent gating mechanisms. This algorithm strikes a balance between floating-point operations (FLOPs) and parallelism, enabling the use of half-precision matrix multiplications to leverage modern GPU tensor cores. GLA exhibits competitive performance on language modeling tasks, demonstrating that gated linear attention Transformers can compete with strong baselines while ensuring computational efficiency.

\section{Detail for \textsc{mqar} and its vanriants} 
\label{appendix: ana_exp}

This section provides additional details for the analysis experiments discussed in Sec.~\ref{sec:3}.

\subsection{General Data Configuration}
We utilize mixed input sequence lengths and varying key-value pair counts for the \textsc{MQAR} tasks. The parameters \( L \) (sequence length), \( N \) (number of key-value pairs), and \( \alpha \) are employed to adjust these properties. Unless specified otherwise, the vocabulary size is set to 20,000. The training set consists of sequence lengths \( L \in \{64, 128, 256, 512\} \), with the corresponding number of key-value pairs \( N \) satisfying \( 2^N \leq \frac{L}{2} \). For empty positions, we insert random values as padding. For each configuration, there are 20,000 examples in the training set and 100 examples in both the validation and test sets. All other settings are consistent with the original \textsc{mqar} task described in \citep{arora2024zoology}.

\subsection{Positional Pattern Change}
We categorize position patterns into three distinct types: standard \textsc{mqar}, last, and shuffle. For the standard \textsc{mqar} pattern, we follow \citeauthor{arora2024zoology}, where the key-value pairs are clustered at the beginning of the sequence, and queries are randomly scattered in the remaining positions. 
In the last pattern, we divide the input sequence into two equal-length segments, placing all key-value pairs in the last portion of the first segment, while queries are randomly distributed in the second segment. 
For the shuffle pattern, we again split the entire input sequence into two equal parts, but we randomly place key-value pairs in the first half, with queries randomly scattered in the second half. 
All other settings remain consistent with the general configuration.

\subsection{N-gram Gathering}
We refer to the original MQAR task as the k1v1 pattern. To assess the model's capability and its tendency to rely on shortcuts in n-gram patterns, we increase the number of tokens in both the keys and values. In this setup, to ensure that the model can effectively distinguish between the key and the value, we introduce special separators between token groups in both the key and value sequences.

Specifically, a key-value pair in the \textit{K2V2} pattern appears as:
\texttt{<SEP> $k_0$ $k_1$ <SEP\_KV> $v_0$ $v_1$ <SEP>}
Although the addition of special symbols may increase the length of individual key-value pairs and cause the model to focus on these symbols, it is necessary. Without the special symbols, a key with \( N \) tokens would be indistinguishable from a key with just 1 token, as the model could rely solely on the last token in the \( N \)-token key to retrieve the corresponding value. Similarly, cases involving \( M \)-token values could often be reduced to multiple instances of \( M-1 \) 2-gram patterns, which would fail to verify if the model truly follows the n-gram relationship. Thus, including these special characters is crucial.

It is also important to note that when the value consists of more than one token, accuracy is measured based on the entire value sequence. In other words, all \( M \) tokens of the value must be predicted correctly for the key-value pair to be considered correct.

\subsection{Noise Injection}

We test the robustness of the two previously mentioned shortcut patterns.

For position robustness, we divide the input sequence into two equal-length segments. Then, we further split the first half into \( N \) sections by position. In our experiments, we set \( N = 4 \). In each of these \( N \) sections, we insert the same key, each corresponding to a different value. The same key is then placed as a query in the second half of the sequence. The model is considered correct if it recalls any of the corresponding values. This setup allows us to assess whether the model, trained on different position patterns, exhibits any positional bias when recalling key-value pairs—specifically, whether it tends to favor key-value pairs from certain positions.

In this configuration, the test data consists of sequence lengths \( L \in \{64, 128, 256\} \) and corresponding key-value pairs \( N \in \{8, 16, 32\} \), with 100 examples per configuration.

For n-gram robustness, we introduce noisy keys by fixing the last \( n-1 \) tokens of the \( n \)-token key while randomly replacing one token. This creates n-gram noise. Specifically, in the k2v2 setup, we fix the second position in the key, and the number of noisy key-value pairs matches the number of original key-value pairs.

\subsection{Training and Evaluation Configuration}

For all synthetic runs on \textsc{MQAR} and its variants, we apply the following training protocol:

\begin{itemize}
    \item \textbf{Optimizer and Schedule}: We use weight decay of 0.1, a warmup duration of 10\%, and a linear warmup schedule. The AdamW optimizer is employed. For each setting, we sweep the learning rates \( \text{lr} \in \{1\text{e-3}, 5\text{e-4}, 1\text{e-4} \} \). All models are trained for 30 epochs without early stopping with 8$\times$A100-40GB GPU.
    
    \item \textbf{Data}: Each model is trained and evaluated on 20,000 training examples and 100 validation examples per setting. The random seed for each configuration is fixed, and the data, as well as its order, remains constant across all runs.
    
    \item \textbf{Loss}: During training, we calculate the cross-entropy loss (\textsc{CELoss}) at all value positions corresponding to each query. For n-gram setups, where special tokens are present, we additionally compute the loss for the last <\texttt{SEP}> token. We found this helps the model understand the role of special tokens in segmenting the sequence.
\end{itemize}

\noindent\textbf{Evaluation}: We evaluate the model using the checkpoint that performs best on the validation set. The training, validation, and test data are constructed in the same way, with identical input sequence lengths and numbers of key-value pairs, differing only in the number of data points and random seeds. However, specific settings are applied for the robustness experiments.

A case is considered correct only if all key-value pairs are predicted correctly.
In multi-value setups (\textsc{knvm}), all \( M \) values corresponding to a key must be predicted accurately to be considered correct. 
The final results are reported as the average across all data points.

\section{Detail for Model} 
\label{appendix: Models}

For our experiments, we adopt the standard Mamba-130M configuration from \citep{gu2023mamba}, as smaller models tend to struggle with the \textsc{MQAR} tasks and do not accurately reflect the shortcut phenomenon.
All models are trained on the same subset of the SlimPajama~\citep{cerebras2023slimpajama} dataset with the \textit{GPT-NeoX} tokenizer for downstream tasks.
All models are trained with AdamW~\citep{loshchilov2018decoupled} using a maximum learning rate of 8e-4. All models are trained on 3B tokens with a batch size of about 0.5M tokens. We use a cosine learning rate schedule with a warmup of 10\% steps. We use a weight decay of 0.01, and gradient clipping of 1.0.

The settings for other baselines are as follows:
\begin{itemize}
    \item \textbf{Pythia-133M}: We reduce the number of layers to 8 while keeping other configurations consistent with Pythia-130M~\citep{biderman2023pythia}. This adjustment minimizes the number of parameters to ensure a fair comparison in our analysis.
    \item \textbf{Hyena-153M}: The configuration matches that of \href{https://huggingface.co/Zymrael/hyena-small-150b-tok}{Hyena-153M}.
    \item \textbf{RWKV-153M}: We set $d_{\text{model}}$ to 1024 and the number of layers to 12, maintaining other configurations consistent with \href{https://huggingface.co/BlinkDL/rwkv-4-pile-430m}{RWKV-Pile-430M}.
    \item \textbf{Mamba-130M}: We adopt the standard configuration from \citeauthor{gu2023mamba}.
    \item \textbf{Mamba-\textit{state\_size}}: We modify the \textit{ssm\_state\_size} of the Mamba model, which is originally set to 16 in the 130M configuration, while keeping all other settings consistent with \citeauthor{gu2023mamba}.
    \item \textbf{Mamba-\textit{Global\_Selection}}: We define \( \Delta t \) as the key to Mamba's selectivity. However, it is constrained by the short convolution and linear transformation. To address this, we propose incorporating more global information into \( \Delta t \). Specifically, we introduce a long convolution module to model distant historical context and use its result to gate the original \( \Delta t \). This is formulated as follows:
    \begin{equation}
    \begin{aligned}
    \mathbf{\Delta_t} &= \mathbf{W}_2 \cdot \sigma\left( \mathbf{W}_1 \cdot \text{Conv}_{\text{short}}(\mathbf{X}_t) \right) \\
    &\odot \sigma\left( \text{Conv}_{\text{long}}(\mathbf{X}_t) \right)
    \end{aligned}
    \end{equation}
where \text{Conv} denotes the convolution operation, \(\mathbf{W}\) denotes the linear transformation operations, \(\sigma\) is the nonlinear activation function, specifically the Silu~\citep{ramachandran2017swish} activation function, and \(\odot\) is the element-wise multiplication operation.
We use the result of the long convolution as a gating mechanism to make global selections on the original \(\mathbf{\Delta}\).
The kernel size of the short convolution is set to 4, as defined by \citep{gu2023mamba}, while the kernel size of the long convolution is set to \(\frac{1}{4}\) of the input sequence length. In the analysis experiments, we add a global selection module to every layer of Mamba, as the relatively short input length ensures that the long convolution does not introduce excessive parameters. However, in downstream tasks, we only incorporate global selection parameters in certain layers to maintain fairness in model parameter counts.

\end{itemize}

\section{Detail for downstream tasks} 
\label{appendix: downstream}

All models are trained on the same subset of the SlimPajama~\citep{cerebras2023slimpajama} dataset with the \textit{GPT-NeoX} tokenizer with a context length of 2048 for downstream tasks.

For downstream evaluation, we use the LM evaluation harness from EleutherAI~\citep{eval2021harness}, as done by previous works.
We evaluate the following tasks/datasets that measure language modeling and common sense reasoning capability:
\begin{itemize}
  \item Wikitext~\citep{merity2016pointer}
  \item HellaSwag~\citep{zellers2019hellaswag}
  \item PIQA~\citep{bisk2020piqa}
  \item WinoGrande~\citep{sakaguchi2021winogrande}
\end{itemize}


We report accuracy for WinoGrande and PIQA and accuracy normalized by sequence length for HellaSwag following prior work~\cite{gu2023mamba}.

\begin{figure*}[t]
    \includegraphics[width=\linewidth]{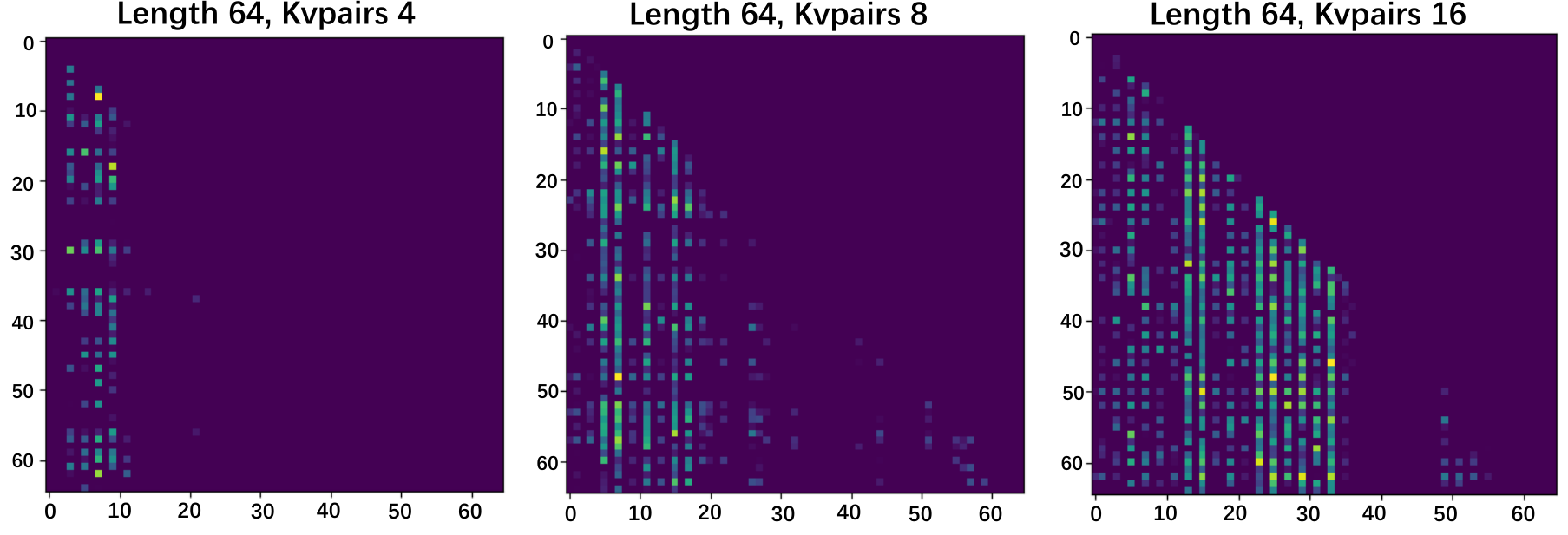}
    \caption{ The attention matrices of Mamba on \textsc{MQAR} tasks, where the input sequence length is 64, and the number of key-value pairs varies from 4 to 16. Lighter colors indicate higher attention weights at specific positions. The results are from the 22st layer of the Mamba model.
}
    \label{fig: mqar_kvpair_attn}
\end{figure*}

\begin{figure*}[t]
    \includegraphics[width=\linewidth]{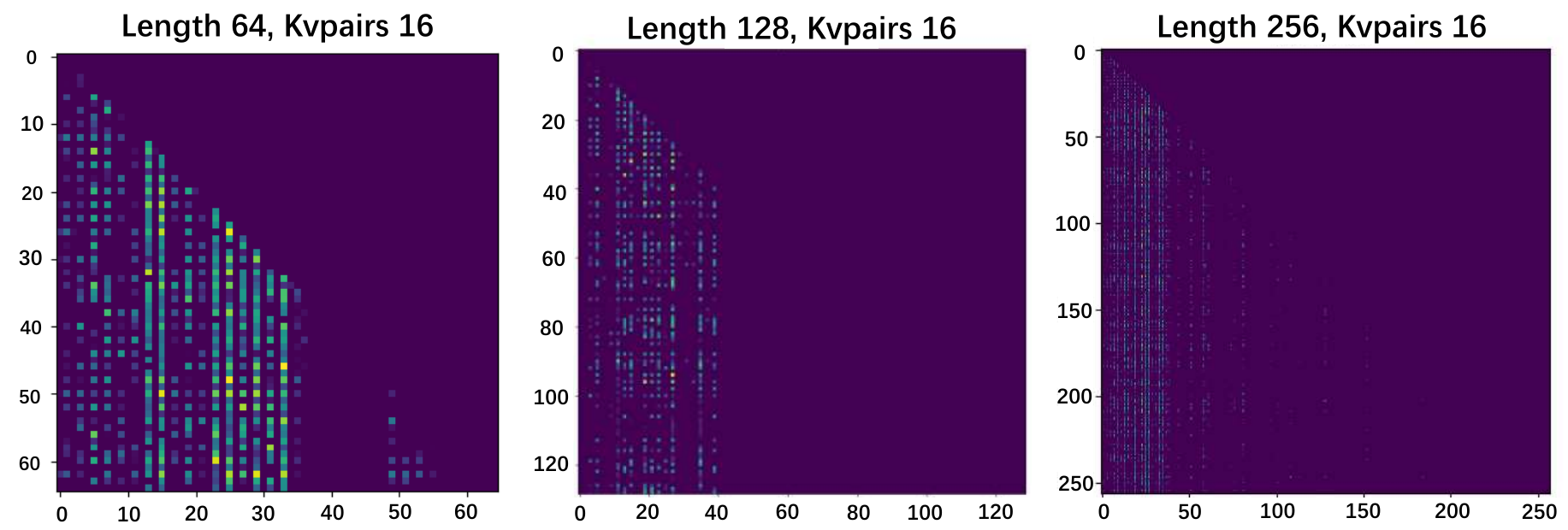}
        \caption{ The attention matrices of Mamba on \textsc{MQAR} tasks, where the number of key-value pairs is 16, and the input sequence length varies from 64 to 256. Lighter colors indicate higher attention weights at specific positions. The results are from the 22nd layer of the Mamba model.
}
    \label{fig: mqar_inputlen_attn}
\end{figure*}

\begin{figure*}[ht]
    \includegraphics[width=\linewidth]{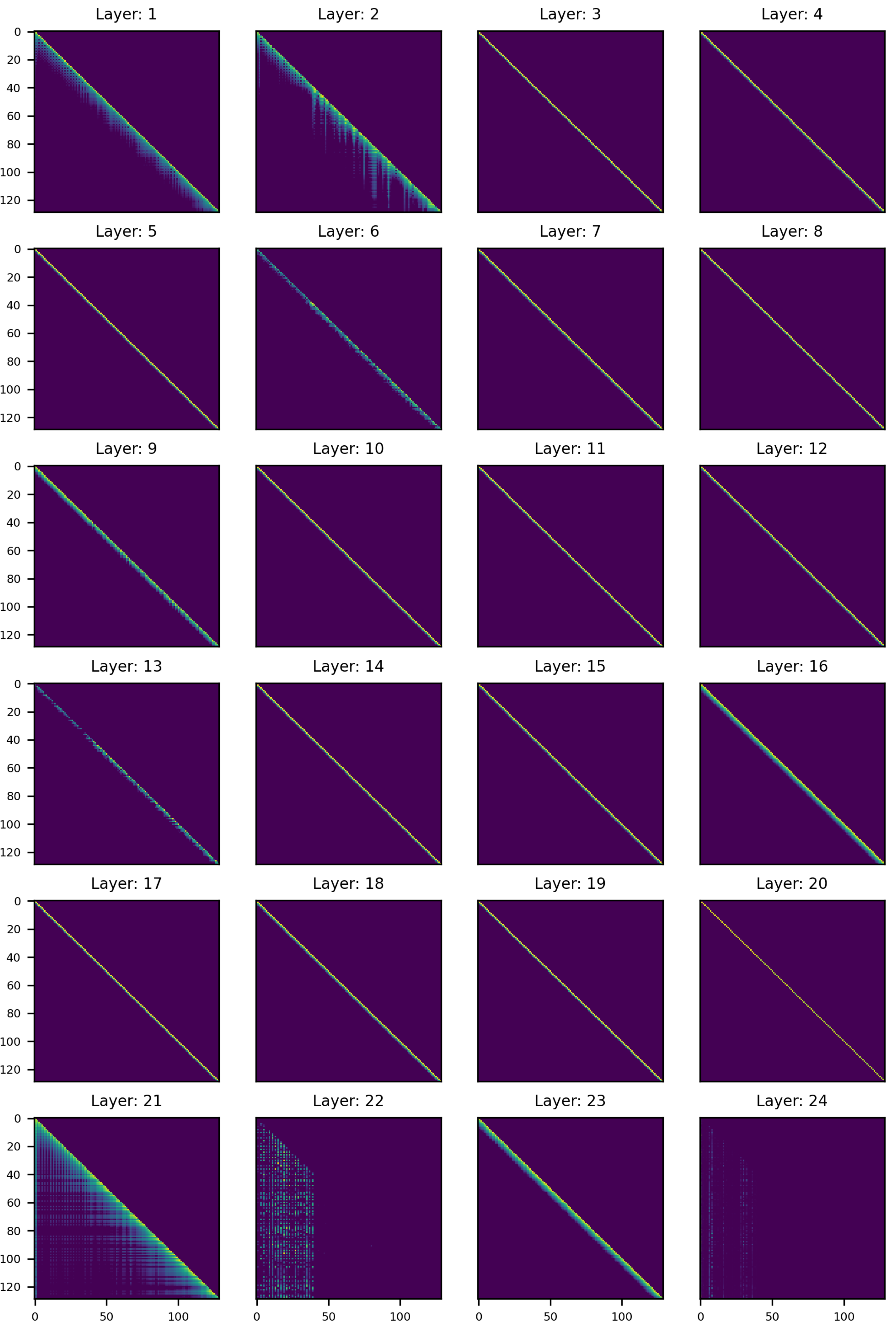}
    \caption{ The attention matrices of all layers of Mamba on \textsc{MQAR} tasks, where the input sequence length is 128, and the number of key-value pairs is 16. Lighter colors indicate higher attention weights at specific positions.
}
    \label{fig: mqar_alllayers}
\end{figure*}

\section{More Analysis on the Selectivity of Mamba} 
\label{appendix: selectivity}

In this section, we provide further analysis to supplement the discussion from the previous section. 

\paragraph{Not all layers are selective:} 
We first examine the attention behavior of each Mamba layer trained on the standard \textsc{MQAR} task and evaluate the model's performance on the same pattern.
The attention matrices for all layers of Mamba on the standard \textsc{MQAR} task are plotted in Figure~\ref{fig: mqar_alllayers}.
One notable observation is that many layers exhibit attention scores concentrated around the main diagonal, while other regions are largely inactive (black).
We consider these layers as primarily responsible for organizing and propagating the hidden state, rather than making specific key-value selections.
In contrast, certain layers demonstrate more pronounced selectivity, such as layer 22 in Figure~\ref{fig: mqar_alllayers}, where the attention distribution sharply focuses on the key-value pair regions.
Clearly, these selective layers are crucial to Mamba's ability to identify relevant information. Therefore, our analysis primarily focuses on these layers.

\paragraph{Perfectly Selective on In-domain Patterns:} 
Mamba performs exceptionally well on the standard pattern of the \textsc{MQAR} task.
We investigate whether Mamba’s perfect performance on these patterns results from its correct selective capabilities.
As shown in Figures~\ref{fig: mqar_kvpair_attn} and~\ref{fig: mqar_inputlen_attn}, these figures depict Mamba's attention distribution under varying input sequence lengths and different numbers of key-value pairs.
It can be observed that, regardless of changes in input length or key-value pair quantity, Mamba consistently captures the correct key-value positions, enabling accurate retrieval of the correct answers.
This observation strongly suggests that the accuracy of the \textsc{MQAR} task is highly dependent on Mamba's ability to capture the correct key-value pairs. 
If the model can accurately pinpoint the positions of these pairs, it has a high probability of correctly retrieving the corresponding value for any given key.

\paragraph{The Role of \(\Delta_t\) in Selectivity:}
In Mamba, the parameters \( A \) and \( B \) play a crucial role in balancing the model's focus between historical information and new inputs.
A key component for this process in common is \(\Delta_t\), which directly governs the model's selective behavior.
By adjusting \(\Delta_t\), Mamba determines whether to prioritize recent inputs or maintain the influence of past states.

The parameter \(\Delta_t\) functions similarly to gating mechanisms in recurrent neural networks (RNNs).
When \(\Delta_t\) is large, the model resets its internal state, focusing on the current input and diminishing the impact of previous information.
In contrast, a small \(\Delta_t\) preserves the existing state, allowing the model to ignore the current input and continue prioritizing historical context.
This behavior parallels the function of gates in RNNs, where a larger gate value highlights new information, and a smaller gate value retains past states.

Mamba can be interpreted as a state-space model (SSM), where \(\Delta_t\) controls the degree of continuity in the system. 
A large \(\Delta_t\) indicates that the model focuses on the current input for an extended period, effectively discarding older information, while a small \(\Delta_t\) implies that the input is transient and has a reduced effect on the model's decision-making process. 
This mechanism enables Mamba to adapt its selective attention dynamically based on the requirements of the task at hand.

\section{Performance on Standard \textsc{Mqar}}
\label{app: zoo_mqar}

\begin{figure*}[t]
    \includegraphics[width=\linewidth]{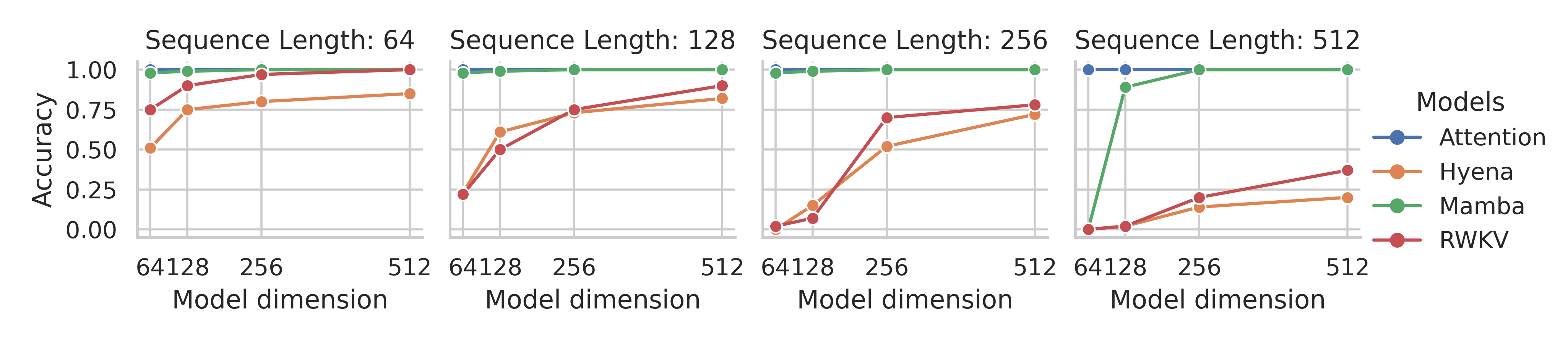}
    \caption{Performance of various models, including H3~\citep{fu2023hungry}, Hyena~\citep{poli2023hyena}, and RWKV~\citep{peng2023rwkv}, alongside \textit{attention-based} models~\citep{touvron2023llama}, across different model dimensions on MQAR tasks. The input sequences vary from 64 to 512 tokens. Other experimental settings are consistent with~\citeauthor{arora2024zoology}.}
    \label{fig: mqar_all_performance}
\end{figure*}

Prior research~\cite {arora2024zoology,arora2024simple} has proven that \textit{attention-based} models surpass \textit{rnn-based} models on \textsc{Mqar} task.
\textsc{Mqar} requires models to memorize key-value pairs in their hidden state, which presents a significant challenge for \textit{rnn-based} models, as they maintain a fixed-size state to handle all historical information.
\ref{fig: mqar_all_performance} depicts performance of different models on the standard \textsc{Mqar} task settings.

It is evident that \textit{rnn-based} models significantly lag behind \textit{attention-based} models, particularly as the length of the input sequence increases.
Attention mechanisms, characterized by their quadratic token interactions, excel in identifying key tokens within the \textsc{Mqar} task and extracting the corresponding values.
Such architectures can utilize the entire past sequence as memory, enabling effective retrieval of prior information and precise recall of associated values.
In contrast, \textit{rnn-based} models are required to store all past information within a single hidden state, relying on this compressed memory to retrieve the relevant keys and values based on the current query.
As the sequence length increases, the compressed historical information grows, making it harder to retrieve the corresponding key-value pairs, leading to worse performance.
An exception to this trend is Mamba, which consistently exhibits performance comparable to attention mechanisms across most settings and significantly outperforms its \textit{rnn-based} counterparts at equivalent model dimensions.
Prior works attribute Mamba's success to its data-dependent features.
By incorporating input-dependent matrices \(\dA\) (via discretization), \(\dB\), and \(\dC\), Mamba effectively retains essential details while discarding irrelevant information.
However, our experiments reveal that this success may not be as intentional or precise as it initially seems.
Mamba's success partly results from its heavy reliance on superficial, local patterns (e.g., attention to the beginning of the input), which we defined as \textbf{"\textit{local pattern shortcuts}"} rather than a genuine ability to recall.

\end{document}